\documentclass[sigconf]{acmart}

\usepackage{times}
\usepackage{soul}
\usepackage{url}
\usepackage{graphicx}
\usepackage{amsmath}
\usepackage{booktabs}
\usepackage{amssymb}
\usepackage{amsthm}
\usepackage{mathrsfs}
\usepackage{subfig}
\usepackage{adjustbox} 
\usepackage[utf8]{inputenc}
\usepackage{multirow}
\usepackage{algorithm}
\usepackage{algpseudocode}
\usepackage{xspace}
\usepackage{amsfonts}
\usepackage{float}
\floatname{algorithm}{Algorithm}
\newcommand{\sysname}{\textit{BayAtk}\xspace}
\usepackage{xcolor}
\newcommand{\revise}[1]{\textcolor{black}{#1}}

\AtBeginDocument{%
  }

\setcopyright{acmlicensed}
\copyrightyear{2018}
\acmYear{2018}
\acmDOI{XXXXXXX.XXXXXXX}

\acmConference[Conference acronym 'XX]{Make sure to enter the correct
  conference title from your rights confirmation emai}{June 03--05,
  2018}{Woodstock, NY}

\acmISBN{978-1-4503-XXXX-X/18/06}

\begin{document}

\title{Transferable Adversarial Examples with Bayesian Approach}

% \author{Anonymous Author(s)}

\author{Mingyuan Fan}
\affiliation{%
  \institution{East China Normal University}
  \city{Shanghai}
  \country{China}}
\email{fmy2660966@gmail.com}

\author{Cen	Chen}
\authornote{Corresponding author.}
\affiliation{%
  \institution{East China Normal University}
  \city{Shanghai}
  \country{China}}
\email{cenchen@dase.ecnu.edu.cn}

\author{Wenmeng	Zhou}
\affiliation{%
  \institution{Alibaba Group}
  \city{Hangzhou}
  \country{China}}
\email{wenmeng.zwm@alibaba-inc.com}

\author{Yinggui Wang}
\affiliation{%
  \institution{Ant Group}
  % \streetaddress{1 Th{\o}rv{\"a}ld Circle}
  \city{Beijing}
  \country{China}}
\email{wyinggui@gmail.com}

\begin{abstract}
    The vulnerability of deep neural networks (DNNs) to \textit{black-box adversarial attacks} is one of the most heated topics in trustworthy AI.
    In such attacks, the attackers operate without any insider knowledge of the model, making the cross-model transferability of adversarial examples critical. 
    Despite the potential for adversarial examples to be effective across various models, it has been observed that adversarial examples that are specifically crafted for a specific model often exhibit poor transferability.
    In this paper, we explore the transferability of adversarial examples via the lens of Bayesian approach.
    Specifically, we leverage Bayesian approach to probe the transferability and then study what constitutes a transferability-promoting prior.
    Following this, we design two concrete transferability-promoting priors, along with an adaptive dynamic weighting strategy for instances sampled from these priors.
    Employing these techniques, we present \sysname.
    Extensive experiments illustrate the significant effectiveness of \sysname in crafting more transferable adversarial examples against both undefended and defended black-box models compared to existing state-of-the-art attacks.
\end{abstract}




\received{20 February 2007}
\received[revised]{12 March 2009}
\received[accepted]{5 June 2009}

\maketitle

\section{Introduction}
\label{intro}
Deep neural networks (DNNs) have achieved remarkable success in a broad spectrum of domains \cite{cnn_review,nlp,recommendation}.
Despite their impressive performance, DNNs are highly vulnerable to adversarial attacks \cite{adv_review,bayes_attack,fan2023trustworthiness}.
Specifically, these attacks manipulate DNNs into making incorrect predictions by employing adversarial examples, generated by imposing carefully-crafted slight perturbations on natural samples.
This inherent vulnerability of DNNs presents a significant challenge for deploying DNNs in security-critical scenarios \cite{adv_review,adv_for_obj}.
Consequently, the identification of DNN vulnerabilities is a top priority, and developing sophisticated adversarial attacks emerges as a cornerstone of research in the AI security area.

There have been substantial advances in the development of adversarial attacks.
These attacks operate under a white-box assumption \cite{FGSM,BIM,PGD}, where attackers can access the target model's details, including specific architectures and parameters.
However, the white-box assumption does not align with typical real-world situations where the internal specifics of the target model remain opaque to attackers, i.e., a black-box scenario.
Hence, launching adversarial attacks in the black-box scenario is more challenging and practical.

There are two primary types of black-box attacks, namely query-based attacks \cite{bayes_attack} and transfer-based attacks \cite{Admix,BSR}.
Query-based attacks entail an avalanche of queries to the target model for estimating the necessary information for crafting adversarial examples (input gradients).
However, this resource-intensive query budget incurs not only considerable expense but also risks detection by model administrators, thereby constraining their practical applicability.
\revise{
In contrast, another approach involves employing a proxy model to generate adversarial examples, known as transfer-based attacks.
The underlying intuition is that if the proxy model shares certain similarities with the target model, the adversarial examples generated by the proxy model may also transfer successfully to the target model.
}

The effectiveness of transfer-based attacks largely relies on the transferability of adversarial examples.
Unfortunately, empirical examinations \cite{DIM,PAM} indicate that such examples often present limited transferability, sparking considerable research interest in enhancing transferability.
Intuitively, the concept of transferability can be likened to the generalization ability of models.
Generalization refers to a model's capability to accurately classify unseen data, whereas transferability pertains to the adversarial examples' effectiveness in deceiving unknown target models. Drawing from this analogy, improving transferability can be viewed through the lens of enhancing generalization.
One well-established method to boost generalization is the Bayesian approach, which incorporates prior knowledge into model parameters to alleviate overfitting issues \cite{bayes_stat,bayes_regular}.
However, the field lacks investigations into studying transferability via the lens of the Bayesian approach, leaving a research gap.

To bridge this gap, this paper represents the first attempt to leverage the Bayesian approach to study transferability.
Our investigation aims to answer three fundamental questions: first, how to integrate the transferability of adversarial examples within the Bayesian framework; second, the essence of the Bayesian approach lies in prior selection, yet it remains unclear what constitutes an appropriate prior for the transferability of adversarial examples; and third, after identifying such priors, how to develop suitable priors.
Our responses to these questions lay the foundation for future research seeking to merge the Bayesian approach with the transferability of adversarial examples:
\begin{itemize}
    \item To tackle the first question, we first revisit the conventional approach of generating adversarial examples on the proxy model. This re-examination elucidates that the vanilla transfer-based attack can be interpreted as a Maximum Likelihood Estimation (MLE) problem, under the assumption of a constant prior distribution.
    While MLE can craft optimal adversarial examples for a specific proxy model, it does not account for the performance of such adversarial examples against different models. Thus, we embrace the Bayesian approach by substituting the MLE's prior with more suitable priors.
    \item Regarding the second question, we base our definition of a suitable prior on the decision-making process of DNNs. Specifically, DNNs work by extracting and analyzing features contained in images, and poor transferability is often due to the excessive concentration on corrupting proxy-model-specific features. In response, we define transferability-promoting prior as one that enables the corruption of a broader range of features.
    \item To answer the third question, we develop pixel-level removal prior and region-based soft removal prior. We demonstrate that both priors qualify as transferability-promoting prior. Furthermore, recognizing that the amount of feature information may vary in instances drawn from the priors, we introduce an adaptive dynamic weighting strategy to mitigate.
\end{itemize}

Building on these advances, we introduce our attack method termed as \sysname, which generates adversarial examples by solving an optimization problem associated with the proposed priors and the adaptive dynamic weighting strategy.
The extensive experiments in the benchmark dataset ImageNet show the superior transferability of adversarial examples generated by \sysname over state-of-the-art transfer-based attack methods.
Moreover, the adversarial examples crafted by \sysname also present considerable effectiveness in misleading the physical-world large multi-modal model Claude3.
To summarize, our contributions are threefold:
\begin{itemize}
    \item We are the first to leverage the Bayesian approach to investigate the transferability of adversarial examples. Our answers to the first and second questions not only bridge a significant research gap but also pave the way for a novel line that intertwines the Bayesian approach with the transferability of adversarial examples.
    \item We develop pixel-level removal prior and region-based soft removal prior and demonstrate their effectiveness as transferability-promoting prior.
    Moreover, we design an adaptive dynamic weighting strategy to mitigate the issue of varying feature information amongst different instances sampled from priors.
    \item We conduct extensive experiments on the large-scale dataset ImageNet and Claude3 to examine the performance of \sysname. The results show that \sysname generates more transferable adversarial examples compared to state-of-the-art attacks.
\end{itemize}

\section{Background \& Related Work}
\label{related_works}
\subsection{Adversarial Attack}

Adversarial attacks impose human-imperceptible adversarial perturbation into natural samples to fool the target model and can be roughly categorized into two types: white-box attacks and black-box attacks.
White-box attacks assume that the attackers have full knowledge of the architecture and parameters of the target model; in contrast, attackers are not allowed to access the internal information about the target model in the black-box scenario.

\textbf{White-box Attacks.}
White-box attacks \cite{FGSM,BIM,PGD} typically infuse inputs with the sign of the input gradient to generate adversarial examples.
One of the most prominent techniques among white-box attacks is the Fast Gradient Sign Method (FGSM) \cite{FGSM}, which crafts adversarial examples through a single-step adjustment aligned with the gradient of the input's loss function.
Let $x$ denote the input image and $y$ its ground-truth label.
Given the target model $\mathcal{T}$ and loss function $\mathcal{L}(\cdot, \cdot)$, FGSM crafts an adversarial example $x_{\text{adv}}$ for $x$ as follows:
\begin{equation}
\label{eq_fgsm}
    x_{adv} = x + \alpha \cdot \text{sign} (\nabla_{x} \mathcal{L}(\mathcal{T}(x), y)),
\end{equation}
where $\alpha$ is the step size (typically set to $\alpha=\frac{1}{255}$) and $\text{sign}(\cdot)$ is element-wise sign operation.
FGSM just needs a single forward and backward propagation and thus is computationally efficient.
Building on FGSM, basic iterative method (BIM) \cite{BIM} refines the adversarial example generation process by applying the FGSM technique iteratively with small step sizes.
BIM is formulated as follows:
\begin{equation}
\label{eq_bim}
\begin{split}
    x_{t+1} &= \Pi_{x, \epsilon} \{ \ x_{t} + \alpha \cdot \text{sign} (\nabla_{x_t} \mathcal{L}(\mathcal{T}(x_t), y)) \ \},\\
    &t=0,\cdots,T-1, \ x_0 = x, \ x_{\text{adv}}=x_T,
\end{split}
\end{equation}
where $x_t$ is the adversarial example obtained after the $t$-th iteration, $\epsilon$ is the perturbation magnitude, and $\Pi_{x+\epsilon}$ projects inputs back into a permissible range centered around $x$ with radius $\epsilon$.
Projected gradient descent (PGD) \cite{PGD} is like an extension of BIM but with a random start, i.e., $x_0 \sim \mathcal{U}(x - \epsilon, x + \epsilon)$ where $\mathcal{U}(a, b)$ denotes a uniform distribution between $a$ and $b$.
The randomness helps in avoiding poor initial points that could lead to less effective adversarial examples.
Although these attacks are designed in the white-box scenario, they commonly serve as a backend component for transfer-based attacks.

\textbf{Black-box Attacks.}
In the black-box scenario, attackers employ two primary strategies: query-based attacks \cite{bayes_attack} and transfer-based attacks \cite{MI,TIM,Admix,SSA}.
We here mainly review transfer-based attacks, as this paper focuses on exploring the transferability of adversarial examples.
Transfer-based attacks commonly train a proxy model $\mathcal{M}$ and then produce adversarial examples with the proxy model using white-box attacks like BIM against the target model.
However, adversarial examples produced purely through white-box attacks often fail to effectively mislead unknown target models \cite{MI,BSR}.
As such, several methods have been proposed to enable more transferable adversarial examples.

From the optimization perspective, momentum attack (MI) \cite{MI} incorporates momentum trick into BIM to stabilize the update direction and escape local optima.
Follow-up works have improved MI by leveraging more advanced optimization methods~\cite{nesterov_adv,VT,path_aug}.

From the model perspective, ensemble attack \cite{ensemble_attack} improves transferability by attacking multiple models at once.
\citet{PI} targeted the feature space of the proxy model to create more transferable adversarial examples.
Other research~\cite{li2020learning,fan2023enhance} explored which model architectures can enable better transferability.

From the input perspective, diverse input (DI) \cite{DIM} resizes and pads input images before gradient calculation.
Scale invariant method (SIM) \cite{SIM} leverages the gradient information from images at different scales, and Admix \cite{Admix} introduces a minor fraction of images from alternative categories into the original input image.
Spectrum simulation attack (SSA) \cite{SSA} aims to augment the proxy model in frequency space, which can be approximated by applying a spectrum transformation in inputs.
PAM \cite{PAM} incorporates multiple augmentation pathways to diversify input images.
BSR \cite{BSR} randomly shuffles and rotates the different regions of inputs.
TPA \cite{fantransferability} penalizes gradients of generated adversarial examples with random noises.

\subsection{Adversarial Defense}

The surge in the sophistication of adversarial attacks has driven progress in the development of countermeasures to these threats.
The most straightforward and effective way is adversarial training \cite{adv_review,adv_trained_models}, in which adversarial examples are integrated into the training phase to instruct the model on the correct identification and management of these inputs.
Despite its effectiveness, adversarial training is accompanied by significant training expenses due to the costly nature of generating adversarial examples.

As an alternative, input transformation-based defenses \cite{HGD,RP,RS,JPEG,NRP} present a cost-effective option, which tries to remove adversarial perturbations before feeding them to DNNs.
Basic methods in this category include JPEG compression, total variance minimization, and image quilting.
More sophisticated methods commonly utilize DNNs.
High-level Representation Guided Denoiser (HGD) \cite{HGD} employs a U-Net while Neural Representation Purifier (NRP) \cite{NRP} learns to purify adversarial examples in a self-supervised manner.

\section{The Proposed Attack: \sysname}
\label{approach}
\begin{figure*}[!th]
    \centering
    \includegraphics[width=0.65\textwidth]{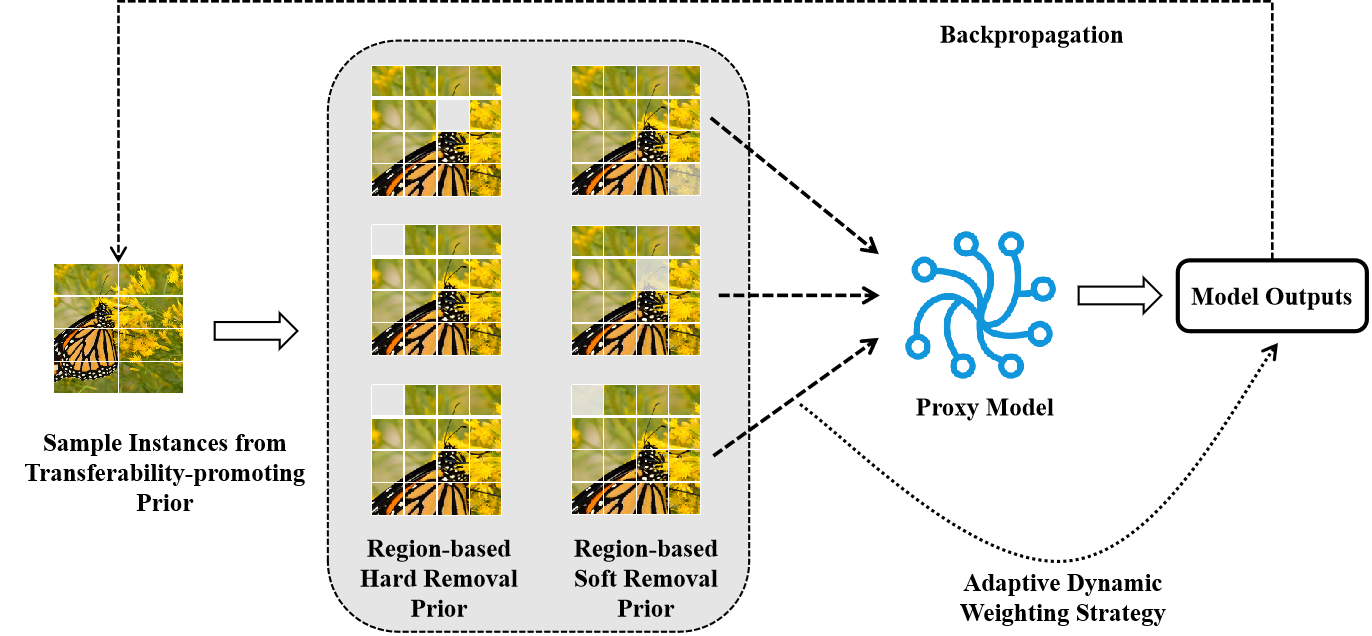}
    \caption{The overview of the proposed attack \sysname. In each iteration, \sysname samples several instances from the transferability-promoting priors and feeds them into the proxy model to obtain the model's prediction probabilities. Then, \sysname weights these probabilities using the adaptive dynamic weighting strategy. Finally, \sysname employs backpropagation algorithm for updates.}
    \label{fig_attack_process}
\end{figure*}

In this section, we develop \sysname, which is outlined in Algorithm \ref{our_algorithm} and Figure \ref{fig_attack_process}.

\subsection{Motivation}

The generation of adversarial examples can be formulated as an optimization task that shares similarities with the training process of DNNs.
The essence of training a DNN lies in identifying a suitable set of parameters that enables the model's high performance on both seen and unseen data.
The primary goal of transfer-based attacks is to produce adversarial examples that are transferable across a diverse range of black-box models.
Therefore, the transferability of adversarial examples is naturally akin to the generalization ability of DNNs.
Recognizing this relationship, it is straightforward to explore model-generalization-promoting techniques to nourish the transferability of adversarial examples.

To this end, we investigate typical model-generalization-promoting techniques \cite{few_shot_learning,cnn_review} and summarize them as follows:
\begin{itemize}
    \item More data. Collecting more data is always a simple yet effective strategy to improve the model's generalization. In the context of adversarial examples, this strategy amounts to training more models with diverse architectures as the proxy models, i.e., ensemble attack \cite{ensemble_attack}. However, despite the simplicity of this strategy, model training is inherently resource-intensive, not to mention the computational costs involved in training multiple models, especially on large-scale datasets like ImageNet.
    \item Robust optimization algorithms. Robust optimization algorithms, like momentum optimization, enjoy the capacity to prevent convergence into suboptimal local minima.
    These algorithms have been extensively examined in existing literature \cite{MI,SIM}.
    \item Model regularizations. This strategy involves restricting the model's expression ability by imposing specific constraints or penalties on its parameters, such as $L_1$ or $L_2$ regularization. These regularizations encourage the model to concentrate on identifying the most discriminative features, thereby mitigating the risk of overfitting. 
    Furthermore, these regularizations embody Bayesian approach, wherein each model parameter is assigned a specific prior distribution, reflecting prior knowledge.
    For example, the adoption of a Gaussian or Laplacian prior leads to the derivation of $L_1$ or $L_2$ regularizations, respectively.
\end{itemize}
Drawing upon the above analysis, Bayesian approach can considerably mitigate the overfitting problem and hence enjoys the potential to promote the cross-model transferability of adversarial examples.
To our best knowledge, there is an absence of research specifically focused on harnessing Bayesian approach to analyze and boost the transferability of adversarial examples, leaving a considerable research gap.
To bridge this gap, we seek to leverage Bayesian approach to study the transferability of adversarial examples.

\begin{algorithm}[!t]
  \caption{\sysname}
  \label{our_algorithm}
  \begin{algorithmic}[1]
    \Require $\mathcal{M}$: the proxy model;
      $x,y$: the natural sample and its ground-truth label;
      $\epsilon$: perturbation budget;
      $\delta$: adversarial perturbation;
      $T$: the number of iterations;
      $\alpha$: step size;
      $Q$: the number of instances to sample per iteration.
      
    \State Adversarial perturbation $\delta$ is randomly initialized within the range between $-\epsilon$ and $\epsilon$.

    \For{each iteration $i = 1$ to $T$}
    
        \State Sample $Q$ instances, denoted as $(x+\delta) \odot m_1, \ldots, (x+\delta) \odot m_Q$ from $p(x+\delta)$. \Comment{Use pixel-level removal prior or region-based soft removal prior.}

        \State Compute the optimization objective $\mathcal{L} = \frac{1}{Q} \sum_{i=1}^Q \mathcal{M}((x+\delta) \odot m_i)[y] \cdot \mathcal{M}((x+\delta) \odot m_i)[y]$.
        \Comment{The first $\mathcal{M}((x+\delta) \odot m_i)[y]$ serves as weights, i.e., our adaptive dynamic weighting strategy.}

        \State Calculate the gradient $g=\nabla_{\delta} \mathcal{L}$

        \State Update adversarial perturbation $\delta = \delta - \alpha \ \text{sign}(g)$.

        \State Clip adversarial perturbation $\delta = \max ( \min(\delta, \epsilon), -\epsilon  )$.
        \Comment{$\max$ and $\min$ are element-wise operations.}

    \EndFor

    \State \textbf{Return:} the crafted adversarial example $x+\delta$.
  \end{algorithmic}
\end{algorithm}

\subsection{Transferable Adversarial Examples with Bayesian Approach}

The adversarial example of $x$ can be written as $x_{\text{adv}} = x + \delta$, where $\delta$ is the adversarial perturbation for $x$.
We formulate the following optimization task for generating adversarial perturbation $\delta$:
\begin{equation}
\label{joint_prob}
\delta^* = \mathop{\arg \min}_{\delta} \ p(x+\delta, y|\mathcal{M}), \ s.t., ||\delta||_{\infty} \leq \epsilon,
\end{equation}
where $p(x+\delta, y|\mathcal{M})$ represents the joint probability of $x+\delta$ and $y$ given the proxy model $\mathcal{M}$, and $\epsilon$ is a constant to control the maximum magnitude of $\delta$, i.e., attack strength.
Intuitively, given $x$ and $y$, Equation \ref{joint_prob} aims to identify the most effective $\delta$ that minimizes the association between $x+\delta$ and $y$.
The above optimization task can be rewritten as follows:
\begin{equation}
\label{our_formulation}
\begin{split}
\delta^* &= \mathop{\arg \min}_{\delta} \ p(x+\delta, y|\mathcal{M}) =  \mathop{\arg \min}_{\delta} \ \frac{p(x+\delta,y,\mathcal{M})}{p(\mathcal{M})} \\
&= \mathop{\arg \min}_{\delta} \ \frac{p(x+\delta,y,\mathcal{M})}{p(x+\delta,M)} \frac{p(x+\delta,\mathcal{M})}{p(\mathcal{M})}\\
&= \mathop{\arg \min}_{\delta} \ p(y|x+\delta, \mathcal{M}) \  p(x+\delta|\mathcal{M}), \\
&s.t., ||\delta||_{\infty} \leq \epsilon,
\end{split}
\end{equation}
where $p(y|x+\delta, \mathcal{M})$ represents the probability that the proxy model $\mathcal{M}$ classifies $x+\delta$ as belonging to class $y$, and $p(x+\delta|\mathcal{M})$ represents the appearing likelihood of $x+\delta$ under model $\mathcal{M}$.
In common transfer-based attacks \cite{BIM,MI}, $p(x+\delta | \mathcal{M})$ is often set to 1, hence focusing primarily on the minimization of $p(y|x+\delta, \mathcal{M})$, i.e., the probability of the model classifying $x+\delta$ as belonging to class $y$.
Unfortunately, this often makes resulting adversarial perturbation $\delta^*$, while potentially optimal for the proxy model $\mathcal{M}$, ineffective when applied to the target model.
This reduction in effectiveness is generally attributed to the overfitting of $\delta^*$ to the proxy model $\mathcal{M}$.
By contrast, Bayesian approach advocates for assigning a specific distribution to $p(x+\delta | \mathcal{M})$ based on prior knowledge, rather than treating $p(x+\delta | \mathcal{M})$ as a constant.
Yet, to our best knowledge, there remains a research gap in understanding what constitutes an appropriate prior for transferable adversarial examples.
We first address this gap, followed by designing our attack.

\begin{figure} [!t]
	\centering
        \raisebox{0.03\height}{
	\subfloat[Original]{
		\includegraphics[scale=0.19]{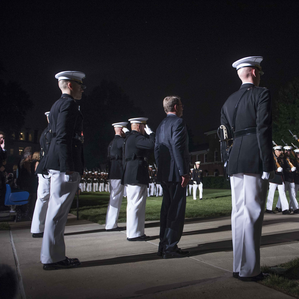}}
        }
	\subfloat[ResNet50]{
		\includegraphics[scale=0.22]{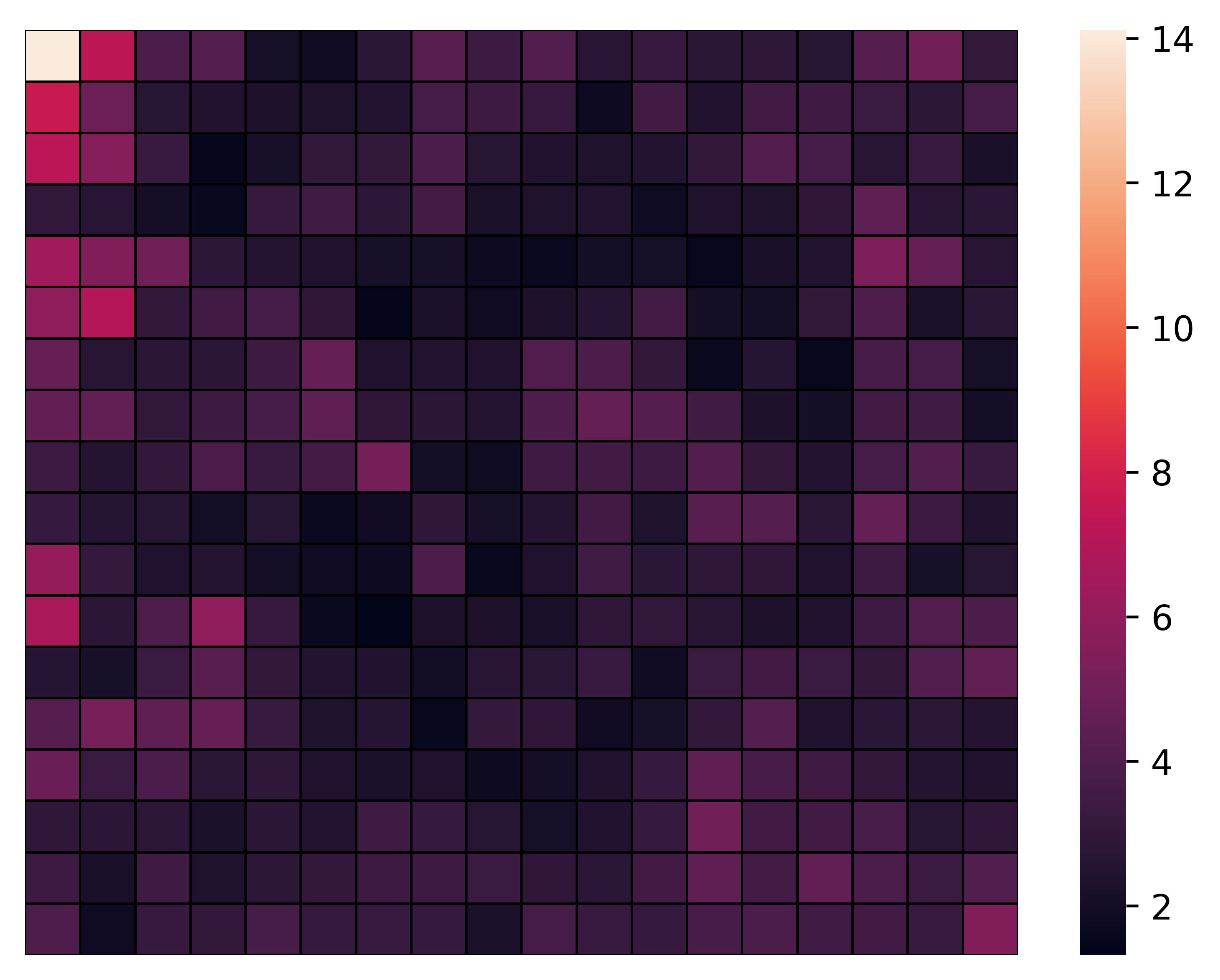}}
	\subfloat[DenseNet121]{
		\includegraphics[scale=0.22]{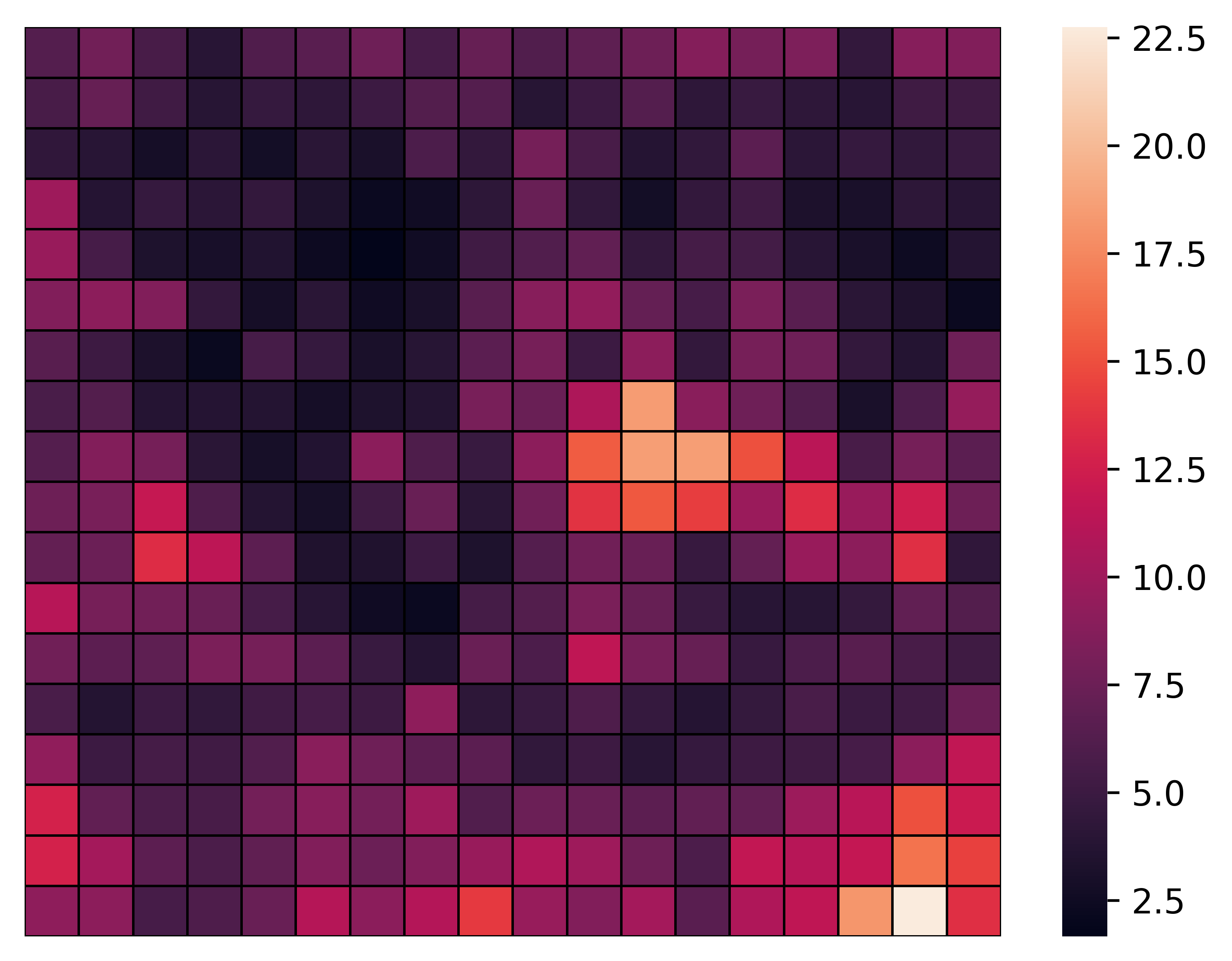}}
	\caption{We divide the leftmost image into different regions and compute the sensitivity of ResNet50 and DenseNet121 to the features in these regions (by summing the absolute values of the gradients). We see that model-specific features learned by ResNet50 and DenseNet121 (the bright squares) are located in different positions, one in the upper left corner and the other in the lower right corner. In contrast, cross-model features are more concentrated in the central area of the image. If we directly maximize the loss of the sample on ResNet50, the generated adversarial example tends to overly focus on disrupting the features in the upper left corner while neglecting the features in the central part of the image.}
	\label{theory_support_fig} 
\end{figure}

\textbf{What constitutes a suitable prior?}
It is widely shared that DNNs classify images by extracting and analyzing features within those images.
Accordingly, the effectiveness of adversarial examples is grounded in the capability of their associated adversarial noises to corrupt these features.
For clarity, we categorize these features as either cross-model features or model-specific features.
Cross-model features are those that are universally learned by various DNNs, whereas model-specific features are exclusive to each model.
This distinction suggests that the limited transferability of adversarial examples can be primarily attributed to their overemphasis on compromising model-specific features while neglecting the cross-model features.
In other words, model-specific features offer a shortcut to fool the proxy model and focusing exclusively on optimizing the probability $p(y|x+\delta, \mathcal{M})$ may excessively exploit this shortcut.
The consequence is the under-emphasis on attacking cross-model features, resulting in adversarial examples that exhibit less transferability.

\revise{
Let us delve deeper into this analysis.
We can describe the probability of the proxy model $\mathcal{M}$ identifying $x$ as $y$ in the following manner:
\begin{equation}
\label{new_form}
% \begin{split}
p(y|x+\delta, \mathcal{M}) = M(x)[y] = \sum_{h_i \in \mathcal{H}} f_i ( h_i^T x ), \ \mathcal{H} = \{h_i\},
% \end{split}
\end{equation}
where $\mathcal{H}$ represents the collection of features $h_i$ learned by the proxy model during its training process.
The term $h_i^T x$ quantifies the relevance of each feature $h_i$ to the input image $x$.
To accommodate the inherent non-linearity of DNNs, we incorporate a nonlinear transformation denoted by $f_i(\cdot)$.
Moreover, $f_i$ is monotonic, as a stronger presence of feature $h_i$ in $x$ typically amplifies $h_i$'s influence on the model's decision-making process.
If $h_i$ positively correlated with class $y$, $f_i ( h_i^T x )$ raises with increasing $h_i^T x$.
Conversely, if negatively correlated, the value decreases.
In either case, $f_i$ remains monotonic in its relationship to $h_i$.
Similarly, we can express the target model $\mathcal{G}$ as $p(y|x+\delta, \mathcal{G}) = \mathcal{G}(x)[y] = \sum_{h_i \in \mathcal{H}} t_i ( h_i^T x )$.
Without loss of generality, let $h_1$ be model-specific features learned by the proxy model, with the remaining features being cross-model.
Moreover, for the sake of discussion, we assume that different features are orthogonal and have equal norms\footnote{This assumption holds when we divide an image into multiple regions, treating each region as a distinct feature. This also serves as one of the reasons for employing the region-based soft removal prior later on.}.
Since $h_1$ is specific to the proxy model, $h_1$ can significantly influences the proxy model's decision-making but has a small impact on the target model's predictions.
Therefore, whether or not input data includes $h_1$ should not significantly affect the target model's predictions, indicating that changes in $t_1$ are small, and $t_1$'s derivative is correspondingly small.
Figure \ref{theory_support_fig} provides a specific example illustrating that $h_1$ does not significantly affect the target model's predictions.
}

\revise{
\citet{trans_cos_sim} demonstrated that the effectiveness of adversarial examples generated by the proxy model against the target model positively correlates with the similarity in the gradient directions of $M(x)[y]$ and $G(x)[y]$.
The gradients of the proxy and target models with respect to input data are given by:
\begin{equation}
\nabla_x M(x)[y] = \sum_{h_i \in \mathcal{H}} f_i' h_i,  \ \nabla_x G(x)[y] = \sum_{h_i \in \mathcal{H}} t_i' h_i.
\end{equation}
The cosine similarity between $f_1' h_1$ and $\nabla_x G(x)[y]$ is as follows:
\begin{equation}
\begin{split}
\text{sim}(f_1' h_1, \nabla_x G(x)[y]) &= \frac{(f_1' h_1) \cdot \left( \sum_{h_i \in \mathcal{H}} t_i' h_i \right)}{\|f_1' h_1\| \left\|\sum_{h_i \in \mathcal{H}} t_i' h_i\right\|} \\
&= \frac{f_1' t_1' \|h_1\|^2}{\|f_1' h_1\| \left\|\sum_{h_i \in \mathcal{H}} t_i' h_i\right\|}.
\end{split}
\end{equation}
We compute the cosine similarity between $\sum_{i \neq 1} f_i' h_1$ and $\nabla_x G(x)[y]$:
\begin{equation}
\begin{split}
	\text{sim}(\sum_{i \neq 1} f_i' h_i, \nabla_x G(x)[y]) &= \frac{\left(\sum_{i \neq 1} f_i' h_i\right) \cdot \left(\sum_{h_i \in \mathcal{H}} t_i' h_i\right)}{\left\|\sum_{i \neq 1} f_i' h_1\right\| \left\|\sum_{h_i \in \mathcal{H}} t_i' h_i\right\|} \\
    &= \frac{\sum_{i \neq 1} f_i' t_i' \|h_i\|^2}{\left\|\sum_{i \neq 1} f_i' h_i\right\| \left\|\sum_{h_i \in \mathcal{H}} t_i' h_i\right\|}.
\end{split}
\end{equation}
Let us consider the relative size of $\text{sim}(\sum_{i \neq 1} f_i' h_i, \nabla_x G(x)[y])$ and $\text{sim}(f_1' h_1, \nabla_x G(x)[y])$:
\begin{equation}
\begin{split}
&\frac{\text{sim}\left(\sum_{i \neq 1} f_i' h_i, \nabla_x G(x)[y]\right)}{\text{sim}(f_1' h_1, \nabla_x G(x)[y])} = \frac{\frac{\sum_{i \neq 1} f_i' t_i' ||h_i||^2}{||\sum_{i \neq 1} f_i' h_i || }}{\frac{f_1' t_1' ||h_1||^2}{||f_1' h_1||}} \\
&= \frac{\sum_{i \neq 1}  f_i' t_i'}{\sum_{i \neq 1} f_i} \frac{f_1}{f_1 t_1} = \frac{\sum_{i \neq 1} f_i \frac{t_i}{t_1}}{\sum_{i \neq 1} f_i}.
\end{split}
\end{equation}
Because $t_i$ is significantly smaller than $t_1$ for $i \neq 1$, it follows that $\text{sim}(\sum_{i \neq 1} f_i' h_i, \nabla_x G(x)[y])$ is greater than $\text{sim}(f_1' h_1, \nabla_x G(x)[y])$.
Put another way, if a prior can facilitate the disruption of cross-model features, it can be considered a transferability-enhancing prior.
In light of the above insight, we define transferability-promoting prior as follows:
}
\begin{definition}
\label{def1}
    (\textbf{Transferability-promoting Prior}) A prior is considered transferability-enhancing if it directs adversarial examples to solely disrupt cross-model features.
\end{definition}

However, due to the black-box nature of the target model and the complexity of DNNs, identifying the cross-model features is indeed intractable, which somewhat undermines the applicability of Definition \ref{def1}.
Therefore, we propose an inclusive version of Definition~\ref{def1}:

\begin{definition}
\label{def2}
    (\textbf{Transferability-promoting Prior, Relaxed Version}) A prior is considered transferability-enhancing if it encourages adversarial examples to corrupt a broader range of features in a more balanced manner.
\end{definition}

In Definition \ref{def2}, the focus shifts from solely targeting cross-model features to disrupting a wider range of features in a balanced manner.
Intuitively, this also can mitigate the excessive attention on corrupting model-specific features during the generation of adversarial examples, thereby promoting the transferability of the crafted adversarial examples.
\revise{
Actually, the intuition of disrupting a broader range of features has been explored in previous literature \cite{hu2022transferability} and our contribution lies in formally validating this intuition and providing an explanation within a Bayesian framework.
}
In the remainder of this paper, unless explicitly stated otherwise, we adopt Definition \ref{def2} by default.

\subsection{Pixel-level Removal Prior}

We here present a transferability-promoting prior which we term the pixel-level removal prior.
Simply put, the pixel-level removal prior randomly nullifies a pixel value from $x$.
Let $x$ be of dimension $K$.
Defined formally, the pixel-level removal prior is expressed as $p(x) = x \odot m_l$, where $m_l$ is a binary vector of the same shape as $x$.
In $m_l$, all elements are set to 1, except for $l$-th element, which is set to 0.
The index $l$ is sampled from a categorical distribution over the set $\{1, 2, \dots, K\}$, with each element having an equal probability of $\frac{1}{K}$ of being selected.

The intuition behind the pixel-level removal prior is straightforward.
Applying the pixel-level removal prior to Equation \ref{our_formulation} intuitively makes the generated adversarial examples that are not only threatening to the proxy model but also less sensitive to the removal of a single-pixel value as well as the features associated with the pixel.
In this way, the pixel-level removal prior could potentially address the issue of adversarial examples being overly reliant on exploiting specific features.
To substantiate this rationale, we engage in a formal examination (Theorem \ref{theorem_1}).

\begin{theorem}
\label{theorem_1}
Pixel-level removal prior is a transferability-promoting prior.
\end{theorem}

\begin{algorithm}[!t]
  \caption{\sysname (Targeted Version)}
  \label{target_algorithm}
  \begin{algorithmic}[1]
    \Require $\mathcal{M}$: the proxy model;
      $x, y_{t}$: the natural sample and target label;
      $\epsilon$: perturbation budget;
      $\delta$: adversarial perturbation;
      $T$: the number of iterations;
      $\alpha$: step size;
      $Q$: the number of instances to sample per iteration.
      
    \State Adversarial perturbation $\delta$ is randomly initialized within the range between $-\epsilon$ and $\epsilon$.

    \For{each iteration $i = 1$ to $T$}
    
        \State Sample $Q$ instances, denoted as $(x+\delta) \odot m_1, \ldots, (x+\delta) \odot m_Q$ from $p(x+\delta)$.

        \State Compute the optimization objective $\mathcal{L} = \frac{1}{Q} \sum_{i=1}^Q \mathcal{M}((x+\delta) \odot m_i)[\neg y_{t}] \cdot \mathcal{M}((x+\delta) \odot m_i)[\neg y_{t}]$.

        \State Calculate the gradient $g=\nabla_{\delta} \mathcal{L}$

        \State Update adversarial perturbation $\delta = \delta - \alpha \ \text{sign}(g)$.

        \State Clip adversarial perturbation $\delta = \max ( \min(\delta, \epsilon), -\epsilon  )$.

    \EndFor

    \State \textbf{Return:} the crafted adversarial example $x+\delta$.
  \end{algorithmic}
\end{algorithm}

We first restate the minimization of Equation \ref{our_formulation} with $p(x+\delta|\mathcal{M})=1$ as follows:
\begin{equation}
\label{new_common_attack}
\begin{split}
\delta^* &= \mathop{\arg \min}_{\delta} \ p(y|x+\delta, \mathcal{M}) =  \mathop{\arg \min}_{\delta} \sum_{h_i \in \mathcal{H}} f_i ( h_i^T (x+\delta) ) \\
&s.t., ||\delta||_{\infty} \leq \epsilon.
\end{split}
\end{equation}
Let us now proceed to the proof of Theorem \ref{theorem_1}.
When leveraging the pixel-level removal prior as $p(x)$, the optimization of $p(y|x,\mathcal{M})p(x)$ translates to:
\begin{equation}
\label{new_our_attack}
\begin{split}
&p(y|x+\delta, \mathcal{M}) \ p(x+\delta) = \mathbb{E}_{p(x+\delta)} \{ p(y|x+\delta, \mathcal{M}) \} \\
&= \frac{1}{K} \sum_{l=1}^K \sum_{h_i \in \mathcal{H}}  f_i ( h_i^T ((x+\delta) \odot m_l) ) \\
&=\frac{1}{K} \sum_{l=1}^K \sum_{h_i \in \mathcal{H}}  f_i ( h_i^T (x+\delta) - h_i[l] \cdot (x+\delta)[l] ) \\
&=  \frac{1}{K} \sum_{l=1}^K \sum_{h_i \in \mathcal{H}}  f_i ( h_i^T (x+\delta) ) \\
&- \frac{1}{K} \sum_{l=1}^K \sum_{h_i \in \mathcal{H}_1}  f_i' ( h_i^T (x+\delta) ) \ (h_i[l] \cdot  (x+\delta)[l]) \\
&=  \sum_{h_i \in \mathcal{H}}  f_i ( h_i^T (x+\delta) ) \\
&- \frac{1}{K} \sum_{h_i \in \mathcal{H}_1}  f_i' ( h_i^T (x+\delta) ) \ \sum_{l=1}^K (h_i[l] \cdot  (x+\delta)[l]) \\
&=  \sum_{h_i \in \mathcal{H}} f_i ( h_i^T (x+\delta) ) - \frac{1}{K}  f_i' ( h_i^T (x+\delta) ) \  h_i^T (x+\delta) \\
&s.t., ||\delta||_{\infty} \leq \epsilon,
\end{split}
\end{equation}
where the operator $(x)[i]$ signifies the extraction of the $i$-th element from the vector $x$.
% The subscript of $m_l$ suggests that the $l$-th element in $m_l$ is set to zero and 
The summation regarding $l$ averages the effect of removing each pixel across all positions indexed by $l$.
The third line to the fourth line in Equation \ref{new_our_attack} hires Taylor expansion.

The last row of Equation \ref{new_our_attack} illustrates that the focal point of optimization becomes more uniformly allocated among features, in contrast with Equation \ref{new_common_attack}.
More precisely, in Equation \ref{new_common_attack}, if $x+\delta$ overly concentrates on compromising feature $h_i$, the gradient magnitude associated with $h_i$ should be considerable, so as to shift optimization attention towards $h_i$.
When coming to Equation \ref{new_our_attack}, $ f_i ( h_i^T (x+\delta) )$ is coupled with a counteracting force term $\frac{1}{K}  f_i' ( h_i^T (x+\delta) ) h_i^T (x+\delta)$\footnote{Based on mathematical principles, as long as $f_i$ is monotonic, it can be derived that the latter term always act as a counteracting force to the preceding term, regardless of the sign of $f_i$ or $h_i^T (x+\delta)$.}.
The magnitude of the countering term is proportionally related to the gradient magnitude associated with $h_i$, thus ensuring that features that are dominantly corrupted in Equation \ref{new_common_attack} bear greater regulation in Equation \ref{new_our_attack}.
Therefore, the pixel-level deletion prior is qualified as a corrective mechanism, encouraging a more balanced corruption across features, i.e., Theorem \ref{theorem_1}.

\begin{table}[!th]
\caption{The attack success rates (\%) on six normally-trained models. The adversarial examples are crafted in Inc-v3, Inc-v4, IncRes-v2, and Res-152, respectively. The best attack results are highlighted in bold. "-" denotes the white-box scenario.}
\label{result_normal_model}
\centering
\scriptsize
\begin{tabular}{@{}cccccccc@{}}
\toprule
Proxy                & Attack & Inc-v3        & Inc-v4        & IncRes-v2     & Res-50        & Res-101       & Res-152       \\ \midrule
\multirow{7}{*}{Inc-v3}    & MI     & -             & 50.3          & 46.6          & 45.0          & 40.7          & 39.6          \\
                           & DI     & -             & 48.1          & 37.4          & 37.7          & 32.0          & 30.2          \\
                           & PI     & -             & 55.4          & 49.0          & 49.1          & 44.1          & 43.4          \\
                           & SSA    & -             & 64.3          & 58.2          & 54.9          & 51.4          & 49.1          \\
                           & PAM    & -             & 74.8          & 75.0          & 67.6          & 68.9          & 62.6          \\
                           & BSR    & -             & 94.3          & 92.9          & 91.5          & 85.5          & 88.7          \\
                           & \sysname   & -             & \textbf{97.0} & \textbf{95.3} & \textbf{95.6} & \textbf{89.7} & \textbf{92.4} \\ \midrule
\multirow{7}{*}{Inc-v4}    & MI     & 61.6          & -             & 45.7          & 46.7          & 42.4          & 41.0          \\
                           & DI     & 53.0          & -             & 35.8          & 31.9          & 28.5          & 30.4          \\
                           & PI     & 59.5          & -             & 45.2          & 49.7          & 41.6          & 43.5          \\
                           & SSA    & 68.5          & -             & 55.8          & 55.5          & 47.6          & 47.2          \\
                           & PAM    & 86.5          & -             & 80.0          & 78.4          & 75.4          & 72.8          \\
                           & BSR    & 93.8          & -             & 91.6          & 97.3          & 87.2          & 87.4          \\
                           & \sysname   & \textbf{96.3} & -             & \textbf{95.0} & \textbf{98.6} & \textbf{90.4} & \textbf{91.9} \\ \midrule
\multirow{7}{*}{IncRes-v2} & MI     & 59.6          & 52.4          & -             & 47.3          & 46.2          & 45.6          \\
                           & DI     & 54.7          & 48.3          & -             & 37.9          & 36.5          & 34.0          \\
                           & PI     & 61.4          & 57.6          & -             & 49.4          & 46.8          & 46.1          \\
                           & SSA    & 75.7          & 67.3          & -             & 58.7          & 58.1          & 54.9          \\
                           & PAM    & 88.0          & 84.7          & -             & 75.9          & 80.1          & 70.5          \\
                           & BSR    & 91.1          & 91.2          & -             & 90.1          & 88.6          & 86.0          \\
                           & \sysname   & \textbf{92.8} & \textbf{95.2} & -             & \textbf{93.6} & \textbf{91.1} & \textbf{91.5} \\ \midrule
\multirow{7}{*}{Res-152}   & MI     & 54.1          & 48.5          & 45.2          & 83.7          & 85.5          & -             \\
                           & DI     & 56.4          & 50.7          & 46.5          & 81.9          & 85.0          & -             \\
                           & PI     & 62.7          & 54.7          & 47.1          & 81.4          & 83.0          & -             \\
                           & SSA    & 65.9          & 61.7          & 56.8          & 91.7          & 94.0          & -             \\
                           & PAM    & 76.2          & 73.4          & 73.8          & 89.6          & 98.5          & -             \\
                           & BSR    & 90.9          & 88.8          & 95.0          & 96.5          & 97.5          & -             \\
                           & \sysname   & \textbf{93.4} & \textbf{91.8} & \textbf{95.8} & \textbf{98.8} & \textbf{98.4} & -             \\ \bottomrule
\end{tabular}
\end{table}

\subsection{Region-based Soft Removal Prior}
\label{approach_region_soft}

In Equation \ref{new_our_attack}, we observe that the effectiveness of the counteracting force term scales linearly with $K$, the total number of pixels of $x$.
The inherently high number of pixels in typical images may lead to a reduction in the intended counteracting effect.
To mitigate this problem, we introduce the region-based soft removal prior.
The core idea of the region-based soft removal prior is straightforward: as shown in Figure \ref{fig_attack_process}, it groups multiple contiguous pixels into a region, analogous to treating a group of pixels as a single entity much like the single pixel in the pixel-level removal prior.
Doing this can significantly decrease the value of $K$, thereby strengthening the influence of the counteracting force term.

Moreover, the region-based soft removal prior does not make the outright elimination of a region; instead, it employs an element-wise scaling operation on all elements within a region.
This soft scaling serves as a generalized extension of the removal operation, thus it holds the potential to confer additional advantages (See Section \ref{ablation_study} for validation).

In practice, we partition $x$ into a grid of non-overlapping regions, i.e., chessboard shape.
Then, through random selection, we choose one region to undergo the element-wise scaling operation.
To formalize, the partitioned $x$ along with $m$ are denoted as $x = [B_1,\cdots, B_d]$ and $m = [m_1,\cdots,m_d]$, respectively.
Each $m_i$ starts with a value of $1$, and one randomly picked $m_i$ is filled with a random scalar between $0$ and $1$.

Notice that the above process amounts to sampling from a joint probability density function: one sample from $[B_1,\cdots, B_d]$ and another from a uniform distribution over $[0, 1]$.
These sampling procedures are independent, ensuring that the region-based soft removal prior adheres to probabilistic principles.

\subsection{Adaptive Dynamic Weighting Strategy}

The instances drawn from $p(x+\delta)$ hold varying amounts of feature information.
For example, a complete removal of a region leads to a more substantial loss of feature information as compared to a soft removal.
Attacking a less-information instance would yield less effective adversarial examples.
Consequently, the information-rich instances should be endowed with more significant attention during the optimization process.

To achieve this, it is necessary to incorporate a weighting mechanism during the optimization process that discriminatively adjusts according to the amount of feature information present in different sampled instances.
Intuitively, information-rich instances ought to be more distinguishable by the model, typically reflected through higher classification confidence.
Motivated by this understanding, we adjust the weighting of the sampled instances in the optimization process in direct proportion to the model's confidence in accurately classifying them as $y$.
In this way, instances contributing valuable features are prioritized, thereby potentially leading to more transferable adversarial examples.
Formally, the final loss function of \sysname is defined as follows:
\begin{equation}
\label{final_loss}
\begin{split}
\delta^* &= \mathop{\arg \min}_{\delta} \ p(y|x+\delta, \mathcal{M}) p(y|x+\delta, \mathcal{M}) \  p(x+\delta|\mathcal{M}) \\
&= \mathcal{M}((x+\delta) \odot m_i)[y] \cdot \mathcal{M}((x+\delta) \odot m_i)[y], \\
&s.t., ||\delta||_{\infty} \leq \epsilon.
\end{split}
\end{equation}
In Equation \ref{final_loss}, the first $p(y|x+\delta, \mathcal{M})$ (or $\mathcal{M}((x+\delta) \odot m_i)[y]$) represents the weighting factor which is not involved in the gradient computation during backpropagation and $i$ is random integer between 1 and $d$.

\subsection{Targeted Version of \sysname}

Algorithm \ref{our_algorithm} provides a description of the non-targeted version of \sysname, whose objective is to cause misclassification without directing the model towards a specific erroneous prediction.
In contrast, targeted attacks take this a step further.
Their goal is not just to make the model misclassify adversarial examples but to force the model to predict a specific, predetermined incorrect class.
Targeted attack is more challenging to execute because they require fine-tuning the adversarial input in such a way that the model consistently arrives at the chosen, incorrect label.
To make \sysname more versatile, we here develop the targeted version of \sysname.

To extend \sysname to targeted attack scenario, we need to make some adjustments to Equation \ref{final_loss}.
The goal of targeted attack is to ensure that the model classifies $x + \delta$ into the desired target label.
In other words, we aim to maximize the probability that the model assigns the target label to $x + \delta$.
Intuitively, this is equivalent to minimizing the sum of the predicted probabilities for all labels other than the target label.
By reducing the likelihood that the adversarial example is classified into any label except the target, we can increase the probability of it being assigned to the target label.
Let target label be denoted as $y_t$.
The loss function for the target version of \sysname can thus be formulated as:
\begin{equation}
\label{final_loss_2}
\begin{split}
\delta^* &= \mathop{\arg \min}_{\delta} \ p(\neg y_{t}|x+\delta, \mathcal{M}) p( \neg y_{t} |x+\delta, \mathcal{M}) \  p(x+\delta|\mathcal{M}) \\
&= \mathcal{M}((x+\delta) \odot m_i)[\neg y_{t}] \cdot \mathcal{M}((x+\delta) \odot m_i)[\neg y_{t}], \\
&s.t., ||\delta||_{\infty} \leq \epsilon.
\end{split}
\end{equation}
In Equation \ref{final_loss_2}, $\neg y_{t}$ represents the set of all labels except the target label, i.e., $\mathcal{Y} - \{y_t,\}$ where $\mathcal{Y}$ is the label space of the model.
Algorithm \ref{target_algorithm} summarizes the targeted version of \sysname.

\section{Experiment}
\label{experiment}

\begin{table}[]
\caption{The attack success rates (\%) on vision transformers. The adversarial examples are crafted in Inc-v3, Inc-v4, IncRes-v2, and Res-152, respectively.}
\label{result_transformer}
\centering
\small
\begin{tabular}{@{}cccccc@{}}
\toprule
Proxy                   & Attack & ViT-B           & PiT-B           & Visformer-S     & Swin          \\ \midrule
\multirow{5}{*}{Inc-v3} & Admix  & 32.9          & 39.2          & 48.1          & 51.7          \\
                              & SSA    & 36.1          & 44.9          & 55.9          & 59.3          \\
                              & PAM    & 38.6          & 49.5          & 61.3          & 63.1          \\
                              & BSR    & 42.4          & 54.8          & 65.9          & 68.3          \\
                              & \sysname   & \textbf{55.6} & \textbf{58.2} & \textbf{69.3} & \textbf{71.2} \\ \midrule
\multirow{5}{*}{Inc-v4} & Admix  & 35.8          & 45.4          & 48.9          & 54.4          \\
                              & SSA    & 41.4          & 46.4          & 57.4          & 61.4          \\
                              & PAM    & 45.0          & 52.2          & 63.9          & 64.8          \\
                              & BSR    & 46.8          & 56.1          & 66.1          & 70.6          \\
                              & \sysname   & \textbf{57.9} & \textbf{60.0} & \textbf{72.6} & \textbf{72.3} \\ \midrule
\multirow{5}{*}{IncRes-v2}    & Admix  & 43.5          & 51.9          & 58.3          & 66.1          \\
                              & SSA    & 49.5          & 54.9          & 64.3          & 74.0          \\
                              & PAM    & 52.6          & 61.2          & 73.4          & 74.5          \\
                              & BSR    & 55.8          & 64.6          & 75.9          & 78.5          \\
                              & \sysname   & \textbf{66.5} & \textbf{69.4} & \textbf{81.1} & \textbf{82.2} \\ \midrule
\multirow{5}{*}{Res-152}   & Admix  & 40.9          & 48.1          & 54.5          & 60.4          \\
                              & SSA    & 44.7          & 49.8          & 61.4          & 69.2          \\
                              & PAM    & 47.9          & 55.5          & 69.7          & 70.2          \\
                              & BSR    & 51.7          & 62.5          & 72.6          & 74.5          \\
                              & \sysname   & \textbf{61.7} & \textbf{65.7} & \textbf{78.8} & \textbf{78.3} \\ \bottomrule
\end{tabular}
\end{table}

\begin{table}[]
\caption{The targeted attack success rates (\%) with the default attack setting.}
\label{result_targeted}
\centering
\scriptsize
\begin{tabular}{@{}cccccccc@{}}
\toprule
Proxy                   & Attack & Inc-v3  & Inc-v4  & IncRes-v2     & Res-50     & Res-101    & Res-152    \\ \midrule
\multirow{3}{*}{Inc-v3} & PAM           & -             & 2.2           & 1.0           & 1.2           & 1.5           & 3.4          \\
                              & BSR           & -             & 3.0           & 2.1           & 2.6           & 2.3           & 3.3          \\
                              & \sysname           & -             & \textbf{4.6}  & \textbf{4.1}  & \textbf{4.5}  & \textbf{4.3}  & \textbf{4.5} \\ \midrule
\multirow{3}{*}{Inc-v4} & PAM           & 2.3           & -             & 0.9           & 1.2           & 1.2           & 3.4          \\
                              & BSR           & 3.1           & -             & 2.3           & 2.7           & 2.1           & 3.1          \\
                              & \sysname           & \textbf{4.9}  & -             & \textbf{4.4}  & \textbf{4.3}  & \textbf{4.6}  & \textbf{4.5} \\ \midrule
\multirow{3}{*}{Res-152}   & PAM           & 8.0           & 5.6           & 6.4           & 9.2           & 14.5          & -            \\
                              & BSR           & 12.4          & 10.1          & 9.5           & 13.4          & 16.5          & -            \\
                              & \sysname           & \textbf{15.7} & \textbf{12.1} & \textbf{11.9} & \textbf{14.1} & \textbf{19.8} & -            \\ \bottomrule
\end{tabular}
\end{table}

\subsection{Setup}

Unless otherwise specified, the settings detailed below adhere to the standard benchmark configuration \cite{SSA,PAM,BSR}.

\noindent \textbf{Models.}
We employ six normally-trained models to serve as our proxy and target models, including Inceptionv3 (Inc-v3), Inceptionv4 (Inc-v4), Inception-Resnet-v2 (IncRes-v2), Resnet-v2-50 (Res-50), Resnet-v2-101 (Res-101), and Resnet-v2-152 (Res-152).
With the emergence of vision models based on transformer architecture, we include four transformer-based visual models, ViT, PiT, Visformer, and Swin, to better evaluate the effectiveness of \sysname.
Additionally, to thoroughly evaluate the effectiveness of \sysname, we also include nine defense models, namely Inc-v3$_{ens3}$, Inc-v3$_{ens4}$, IncRes-v2$_{ens}$ \cite{adv_trained_models}, HGD \cite{HGD}, R\&P \cite{RP}, NIPS-r3 \footnote{https://github.com/anlthms/nips-2017/tree/master/mmd}, JPEG \cite{JPEG}, RS \cite{RS}, and NRP \cite{NRP}.
These defenses cover the previously mentioned two mainstream adversarial defense directions, with the first three using adversarial training and the latter six utilizing input transformation techniques.
Normally-trained models serve as the proxy model throughout the experiment, while normal and defense models are the target models in Section \ref{exp_normal} and Section \ref{exp_defense}, respectively.

\noindent \textbf{Competitor.}
To demonstrate the effectiveness of \sysname, we compare it with various state-of-the-art transfer-based attacks, including MI \cite{MI}, DI \cite{DIM}, PI \cite{PI}, SSA \cite{SSA}, PAM \cite{PAM}, and BSR \cite{BSR}.

\noindent \textbf{Dataset.}
We conduct evaluations in a subset of ImageNet \cite{SSA}, consisting of 1000 images that are well classified by the models used in this paper.
We also include CIFAR-10 to assess the generalizability of \sysname across different datasets.

\noindent \textbf{Evaluation metric.}
We adopt attack success rate (ASR) as the evaluation metric.
ASR measures the ratio of proxy-model-generated adversarial examples that cause misclassification from target models.
A higher ASR indicates better attack performance, i.e., higher transferability of adversarial examples.

\noindent \textbf{Hyperparameters.}
For baselines, we follow the default hyperparameters specified in their original papers.
In line with existing transfer-based attacks \cite{SSA,PAM,BSR}, we use perturbation budget $\epsilon=\frac{16}{255}$, iteration number of $10$, and step size of $\frac{1.6}{255}$ for all attacks.
For \sysname, we partition a single image into $14 \times 14$ small squares, mimicking a $14 \times 14$ chessboard.
In each iteration, \sysname softly removes one of these squares and samples 20 instances ($Q=20$), i.e., the region-based soft removal prior.
We study the influence of sub-components and hyperparameters (the size of chessboard) on the performance of \sysname in Section \ref{ablation_study}.

\subsection{Attack Performance on Normal Models}
\label{exp_normal}

\textbf{Convolution-based models.}
We first evaluate the performance of \sysname on normally-trained models.
To do this, we generate adversarial examples using four proxy models and subsequently evaluate their effectiveness across a variety of target models.
Table \ref{result_normal_model} reports the performance of different attacks over diverse proxy-target model pairs.
Overall, regardless of the specific proxy and target model pairing, \sysname consistently boosts the transferability of adversarial examples across nearly all proxy models by a clear margin.
This consistency suggests that the effectiveness of \sysname is not dependent on particular proxy models being used.
More impressively, the average ASR of \sysname across all tested models stands at approximately 93\%, underscoring \sysname's effectiveness in generating transferable adversarial examples.
This performance is particularly striking when we consider the varying degrees of success across different attack types, such as MI, DI, PI, SSA, PAM, and BSR.
Diving deeper into the individual results, when employing Inception-v3 as the proxy model, the ASR of \sysname surpasses that of the state-of-the-art attack BSR by approximately 3\% $\sim$ 4\%.
These attack results demonstrate \sysname's superior ability in generating transferable adversarial examples.

\begin{table}[]
\small
\caption{The targeted attack success rates (\%). The adversarial examples are crafted in an ensemble of white-box proxy models with iteration number of 500 and logit loss function.}
\label{improved_target}
\begin{tabular}{@{}cccccc@{}}
\toprule
Attack & Inc-v3 & Inc-v4 & IncRes-v2 & ViT-B & PiT-B \\ \midrule
PAM + Logit    & 17.2   & 13.4   & 10.7      & 5.4   & 6.3   \\
BSR + Logit    & 22.2   & 18.2   & 12.9      & 6.6   & 7.6   \\
\sysname + Logit   & \textbf{29.8}   & \textbf{27.1}   & \textbf{22.0}      & \textbf{14.6}  & \textbf{15.6}  \\ \bottomrule
\end{tabular}
\end{table}

\begin{table}[]
\small
\caption{The attack success rates (\%) on three models trained in CIFAR-10. We use Res-50 trained in CIFAR-10 as the proxy model.}
\label{eval_cifar10}
\begin{tabular}{@{}cccc@{}}
\toprule
Attack & Inc-v3 & Inc-v4 & Res-101 \\ \midrule
PAM    & 82.89  & 79.15  & 85.68   \\
BSR    & 85.64  & 82.30  & 88.26   \\
\sysname   & \textbf{89.32}  & \textbf{86.32}  & \textbf{92.82}   \\ \bottomrule
\end{tabular}
\end{table}

\begin{table}[!t]
\caption{The attack success rates (\%) on three adversarially-trained models. The adversarial examples are crafted in Inc-v3, Inc-v4, IncRes-v2, and Res-152, respectively.}
\label{result_defense_model}
\centering
 \small
\begin{tabular}{@{}ccccc@{}}
\toprule
Proxy                & Attack & Inc-v3$_{ens3}$ & Inc-v3$_{ens4}$ & IncRes-v2$_{ens}$  \\ \midrule
\multirow{5}{*}{Inc-v3}    & Admix &  35.4       & 34.0       & 25.9 \\
                           & SSA    & 38.3       & 37.9       & 27.1 \\
                           & PAM    & 39.0       & 38.8       & 28.0 \\
                           & BSR    & 51.0       & 46.6       & 29.1 \\
                           & \sysname    & \textbf{55.6} & \textbf{52.2} & \textbf{32.5} \\ \midrule
\multirow{5}{*}{Inc-v4}    & Admix  & 45.2       & 44.3       & 30.1 \\
                           & SSA    & 47.7       & 45.9       & 31.7 \\
                           & PAM    & 55.4       & 50.5       & 33.2 \\
                           & BSR    & 57.6       & 52.1       & 34.3 \\
                           & \sysname    & \textbf{60.5} & \textbf{56.8} & \textbf{37.8} \\ \midrule
\multirow{5}{*}{IncRes-v2} & Admix  & 52.6       & 49.9       & 49.5 \\
                           & SSA    & 55.3       & 52.0       & 50.2 \\
                           & PAM    & 66.0       & 58.3       & 51.0 \\
                           & BSR    & 71.4       & 63.1       & 51.0 \\
                           & \sysname    & \textbf{75.4} & \textbf{66.7} & \textbf{52.6}  \\ \midrule
\multirow{5}{*}{Res-152}   & Admix  & 44.2       & 34.7       & 27.7 \\
                           & SSA    & 45.4       & 37.0       & 31.7 \\
                           & PAM    & 51.2       & 46.3       & 35.2 \\
                           & BSR    & 78.7       & 74.7       & 51.6 \\
                           & \sysname    & \textbf{82.3} & \textbf{77.8} & \textbf{55.1} \\ \bottomrule
\end{tabular}
\end{table}

\begin{table}[!t]
\caption{The attack success rates (\%) on other defenses. The adversarial examples are crafted in Inc-v3, Inc-v4, IncRes-v2, and Res-152, respectively.}
\label{result_other_defense}
\centering
\small
\begin{tabular}{@{}cccccccc@{}}
\toprule
Proxy                & Attack & HGD  & R\&P & NIPS-r3 & JPEG & RS   & NRP  \\ \midrule
\multirow{5}{*}{Inc-v3}    & Admix  & 30.2 & 28.0 & 30.3    & 40.2 & 33.2 & 21.9 \\
                           & SSA    & 37.1 & 33.3 & 36.7    & 41.1 & 33.8 & 26.7 \\
                           & PAM    & 38.7 & 34.5 & 38.3    & 43.0 & 35.0 & 27.5 \\
                           & BSR    & 49.2 & 43.3 & 48.6    & 53.0 & 44.7 & 35.5 \\
                           & \sysname    & \textbf{53.6} & \textbf{49.9} & \textbf{53.1} & \textbf{58.4} & \textbf{51.4} & \textbf{40.8} \\ \midrule
\multirow{5}{*}{Inc-v4}    & Admix  & 53.2 & 47.8 & 51.0    & 59.4 & 48.2 & 34.4 \\
                           & SSA    & 54.8 & 51.8 & 55.2    & 60.1 & 52.4 & 39.0 \\
                           & PAM    & 60.4 & 57.6 & 61.7    & 67.6 & 58.5 & 43.4 \\
                           & BSR    & 66.4 & 62.3 & 66.6    & 73.3 & 63.1 & 46.9 \\
                           & \sysname    & \textbf{71.4} & \textbf{66.3} & \textbf{73.3} & \textbf{77.0} & \textbf{66.4} & \textbf{52.5} \\ \midrule
\multirow{5}{*}{IncRes-v2} & Admix  & 79.5 & 79.3 & 80.2    & 84.9 & 72.2 & 59.7 \\
                           & SSA    & 82.4 & 80.6 & 82.4    & 86.1 & 74.2 & 61.6 \\
                           & PAM    & 86.7 & 84.6 & 85.2    & 89.8 & 77.6 & 64.5 \\
                           & BSR    & 91.4 & 88.9 & 91.4    & 96.0 & 82.0 & 67.5 \\
                           & \sysname    & \textbf{93.5} & \textbf{93.7} & \textbf{95.9} & \textbf{97.2} & \textbf{85.8} & \textbf{70.8} \\ \midrule
\multirow{5}{*}{Res-152}   & Admix  & 51.4 & 48.1 & 50.7    & 54.0 & 47.6 & 38.1 \\
                           & SSA    & 54.7 & 50.6 & 54.2    & 58.0 & 50.0 & 42.2 \\
                           & PAM    & 62.4 & 56.9 & 60.5    & 66.4 & 56.6 & 47.8 \\
                           & BSR    & 68.3 & 61.9 & 67.8    & 71.8 & 61.4 & 51.9 \\
                           & \sysname    & \textbf{71.8} & \textbf{66.3} & \textbf{70.2} & \textbf{75.7} & \textbf{65.0} & \textbf{57.0} \\ \bottomrule
\end{tabular}
\end{table}

\begin{table}[]
\caption{The oddness scores of BIM and \sysname. The results are averaged over 100 samples.}
\label{oddness_score_eval}
\small
\begin{tabular}{@{}cccc@{}}
\toprule
Proxy       & Res-50 & Res-152 & Inc-v3  \\ \midrule
BIM    & 0.615 & 0.584  & 0.560  \\
\sysname  &\textbf{ 0.282}     & \textbf{0.287} & \textbf{0.322} \\ \bottomrule
\end{tabular}
\end{table}

\begin{table*}[!t]
\caption{Ablation study on different sub-components of \sysname. We report ASRs (\%).}
\label{exp_ablation_study}
\centering
\small
\begin{tabular}{@{}cccc@{}}
\toprule
Strategy                                                                                                 & Inception-v4 & Resnet-50 & Resnet-101 \\ \midrule
BIM                                                                                                      & 21.7 & 15.2 & 13.4      \\
+ \ Pixel-level Removal                                                                                     & 54.5 & 44.8 & 42.2         \\
+ \ Region-based Hard Removal                                                                               & 90.7 & 88.2 & 83.5         \\
+ \ Region-based Soft Removal                                                                               & 95.8 & 94.4 & 88.2         \\
+ \ Region-based Soft Removal \& Adaptive Dynamic Weight Strategy &   97.0     &   95.6     &    89.7     \\ \bottomrule
\end{tabular}
\end{table*}

\begin{table}[!t]
\caption{The ASRs (\%) of \sysname with varying chessboard layouts.}
\label{exp_chess}
\centering
\small
\begin{tabular}{@{}cccc@{}}
\toprule
Chessboard Layout & Inception-v4 & Resnet-50 & Resnet-101 \\ \midrule
7$\times$7       &   95.8     &   94.2     & 87.7        \\
14$\times$14     &   97.0     &   95.6     &    89.7     \\
28$\times$28     &   96.5     &   94.8    &  88.5       \\
56$\times$56     &   94.6     &   92.9    &   86.3      \\ \bottomrule
\end{tabular}
\end{table}

\textbf{Transformer-based models.}
Table \ref{result_transformer} presents the ASRs achieved by different attack methods on four vision transformer models.
The attack results in Table \ref{result_transformer} demonstrate the effectiveness of \sysname across different vision transformer models.
In particular, when using ResNet-152 as the proxy model, \sysname leads with ASRs of 61.7\% for ViT-B, 65.7\% for PiT-B, 78.8\% for Visformer-S, and 78.3\% for Swin.
In contrast, BSR, which performs relatively well, still lags behind \sysname by substantial margins, particularly in the case of ViT-B and Visformer-S, where the differences are most pronounced.

\textbf{Targeted attack results.}
We also evaluate the targeted attack performance of \sysname in comparison to two state-of-the-art transfer-based attack methods, PAM and BSR.
The evaluation setup follows \cite{targeted_attack_nips}, utilizing the same target setting for consistency.
Table \ref{result_targeted} presents targeted ASRs of three attack methods.
The results indicate that \sysname consistently surpasses PAM and BSR across various proxy-target model pairs.
For instance, in Inc-v3, \sysname achieves the highest ASR of 4.6\% for Inc-v4 and 4.5\% for both Res-50 and Res-152, surpassing the results of PAM and BSR.
Similarly, in the Inc-v4, \sysname records an ASR of 4.9\%, which is higher than the other methods.
In Res-152, the superiority of \sysname is even more pronounced, with ASRs reaching 15.7\% compared to 8.0\% for PAM and 12.4\% for BSR against Inception-v3.
This trend highlights the effectiveness of \sysname in executing targeted attacks.

Moreover, we see that the targeted ASRs reported in Table \ref{result_targeted} are somewhat low.
This is primarily because targeted attacks require more attack iterations.
To this end, following \cite{targeted_attack_nips}, we modify the loss function to logit and increase the number of attack iterations to 500.
We then employ an ensemble of white-box proxy models, including ResNet-50, DenseNet121, and VGG16, for re-evaluation.
Table \ref{improved_target} reports the targeted ASRs for various attack methods.
The results indicate a significant improvement in the targeted ASRs across different methods, with \sysname generating more transferable adversarial examples.

\revise{
\textbf{The Evaluation on CIFAR-10.}
We further evaluate the effectiveness of \sysname on CIFAR-10 dataset.
We train Res-50, Res-101, Inc-v3, and Inc-v4 in the training set of CIFAR-10.
ResNet-50 is then used as the proxy model to generate 10000 adversarial examples for the test set of CIFAR-10.
The ASRs of three attack methods are summarized in Table \ref{eval_cifar10}.
We find that \sysname consistently outperforms both PAM and BSR on CIFAR-10 by a significant margin, demonstrating the generalizability of \sysname across different datasets.
}

\subsection{Attack Performance on Defense Models}
\label{exp_defense}

Although many attacks are capable of misleading normally trained models, these attacks might not be as effective against models with defense mechanisms in place.
To further substantiate the superiority of \sysname, we examine the effectiveness of \sysname against defense models.
Table \ref{result_defense_model} and Table \ref{result_other_defense} report the results of different attacks in two mainstream defense paradigms, respectively.

Adversarial training, deemed a gold standard in defense, iteratively exposes models to adversarial examples during the training phase, allowing them to be resilient to similar attacks in the inference stage.
We assess the attack effectiveness of \sysname against three top-tier adversarially-trained models.
Looking at the results, we see a consistent improvement by \sysname over baselines.
Specifically, when employing Inc-v3 as the proxy model, \sysname yields ASRs of 55.6\%, 52.2\%, and 32.5\% against the three adversarially-trained models, respectively.
These results represent a substantial improvement over the best-performing baseline, BSR, which records ASRs of 51.0\%, 46.6\%, and 29.1\%.
\sysname exceeds BSR by margins of 4.6\%, 5.6\%, and 3.4\%.
Furthermore, when evaluating the performance of \sysname with Inception-v4, IncRes-v2, and, Resnet-152 as the proxy models, we observe similar trends.
In short, \sysname is effective for enhancing the cross-model transferability of crafted adversarial examples, in spite of normal or adversarially-trained models.

Input transformation-based defenses, which differ fundamentally from adversarial training, tackle adversarial attacks by applying preprocessing techniques to model inputs with the aim of diminishing the impact of adversarial examples by transforming them back into their benign counterparts.
Overall, as outlined in Table \ref{result_defense_model}, \sysname maintains its superiority when confronted with six state-of-the-art input transformation-based defenses.
While these defenses can strip certain adversarial features, \sysname appears to have a higher capacity to generate adversarial examples that retain their effectiveness after transformation, as evidenced by significantly higher ASRs.
For example, \sysname's ASRs are significantly higher (58.4\% to 77.0\%) compared to BSR (53.0\% to 73.3\%) against JPEG compression defense when leveraging Inception-v3 and Inception-v4 as the proxy models.
When we shift our focus to IncRes-v2 as the proxy model, \sysname achieves the highest ASRs of 93.5\%, 93.7\%, 95.9\%, 97.2\%, 85.8\%, and 70.8\%, significantly surpassing BSR’s performance, which records ASRs of 91.4\%, 88.9\%, 91.4\%, 96.0\%, 82.0\%, and 67.5\%.
The improvements of 2.1\% to 4.3\% across these defenses further illustrate the exceptional capability of \sysname in generating resilient adversarial examples.

\begin{figure} [!t]
	\centering
	\subfloat{
		\includegraphics[scale=0.35]{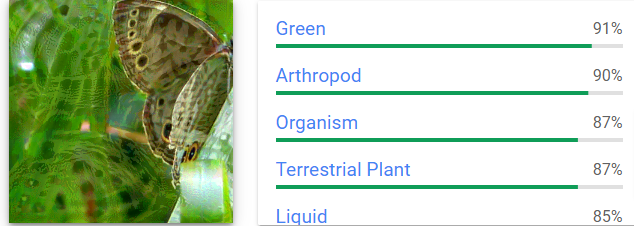}}
    \\
	\subfloat{
		\includegraphics[scale=0.35]{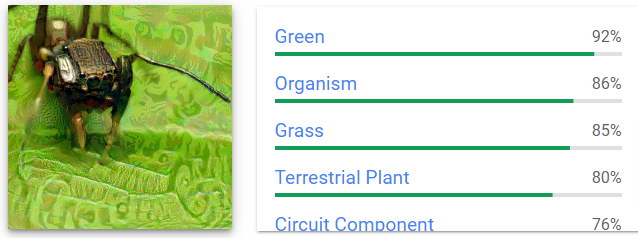}}
    \\
	\subfloat{
		\includegraphics[scale=0.35]{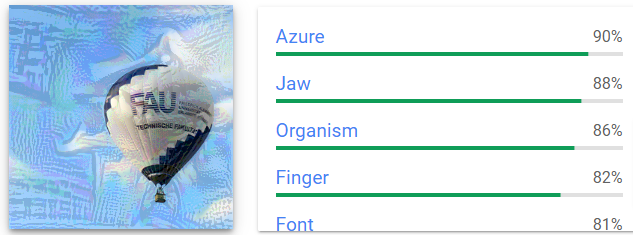}}
	\caption{Three successful attack cases using \sysname. On the left are the adversarial examples generated by \sysname, and on the right are the Top-5 predictions from the Google MLaaS Classification System for these examples.}
	\label{fig_google_attack} 
\end{figure}

\subsection{Oddness Evaluation}

\revise{
We here validate that \sysname can corrupt a wider range of features in a more balanced manner.
We employ BIM and \sysname to generate adversarial examples and then compute the differences between these examples and the original images.
The differences are partitioned into $14 \times 14$ regions, and we calculate the sum of values for each region.
The standard deviation of these sums across different regions is defined as the oddness score.
Intuitively, a lower oddness score suggests that features contained in input samples are disrupted in a more uniform manner.
Table \ref{oddness_score_eval} reports the oddness scores achieved by different methods, averaged over 100 samples.
As observed, \sysname attains a lower oddness score, demonstrating its superior capability to disrupt features in images more uniformly.
}

\subsection{Sub-component Analysis and Sensitivity Analysis}
\label{ablation_study}

In this subsection, we investigate the contributions of different sub-components of \sysname to its performance, alongside examining the influence of chessboard size on its attack effectiveness.

\noindent \textbf{Sub-component analysis.}
Our exploration initiates with BIM, which acts as the foundation attack, i.e., the vanilla transfer-based attack, upon which \sysname and other transfer-based attacks are built.
Table \ref{exp_ablation_study} reports the attack performance of BIM combined with various sub-components.
The effectiveness of BIM alone is observed to be relatively low, with ASRs spanning from 13.4\% to 21.7\%.
This initial performance sets the baseline from which the impact of different sub-components can be quantitatively measured.

Upon integrating pixel-level removal prior to BIM, a significant surge in attack performance is noted, with ASRs jumping to 54.5\%, 44.8\%, and 42.2\%.
This substantial increase validates the effectiveness of pixel-level removal prior as a transferability-promoting prior.
The next evolutionary step replaces the pixel-level removal prior with the region-based hard removal prior.
The performance leap in this step is even more pronounced, propelling ASRs to impressive numbers of 90.7\%, 88.2\%, and 83.5\%.
As discussed in Section \ref{approach_region_soft}, we attribute this significant improvement to its focus on coherent regions rather than isolated pixels, which is key to decreasing the value of $K$ and enhancing the influence of the counteracting force term.
Switching from hard to soft removal yields a further increment in ASRs to 95.8\%, 94.4\%, and 88.2\%, highlighting the benefits of a gentler, perhaps more probabilistic method to region removal.
We speculate that the superior performance of soft over hard removal may stem from its ability to more intricately scale the feature information within images instead of outright elimination.
The final addition of the adaptive dynamic weight strategy together with region-based soft removal pushes the ASRs to their zenith - 97.0\%, 95.6\%, and 89.7\%.
This demonstrates the effectiveness of dynamically adjusting the weight of sampled instances based on the richness of feature information they possess, thereby enhancing the attack performance.

\noindent \textbf{The impact of chessboard layout.}
Table \ref{exp_chess} reports the performance of \sysname with varying chessboard sizes and a $14 \times 14$ chessboard achieves the highest attack effectiveness.
Intuitively, chessboard size that is either excessively large or small detracts from the effectiveness of \sysname.
On the one hand, enlarging chessboard layout, for instance, from $14\times 14$ to $56 \times 56$, results in each grid representing a smaller subset of pixels.
This change suggests a movement away from effectively leveraging the region-based prior towards an emphasis on the pixel-level prior, which, in turn, leads to a degradation in performance of \sysname.
On the other hand, a smaller chessboard implies that an overly large part of $x$ is modified, leading to the elimination of significant portions of image features.
Hence, it appears that the $14 \times 14$ chessboard size likely offers an optimal balance.

\subsection{Towards Efficient \sysname}

\begin{figure}
    \centering
    \includegraphics[width=0.8\linewidth]{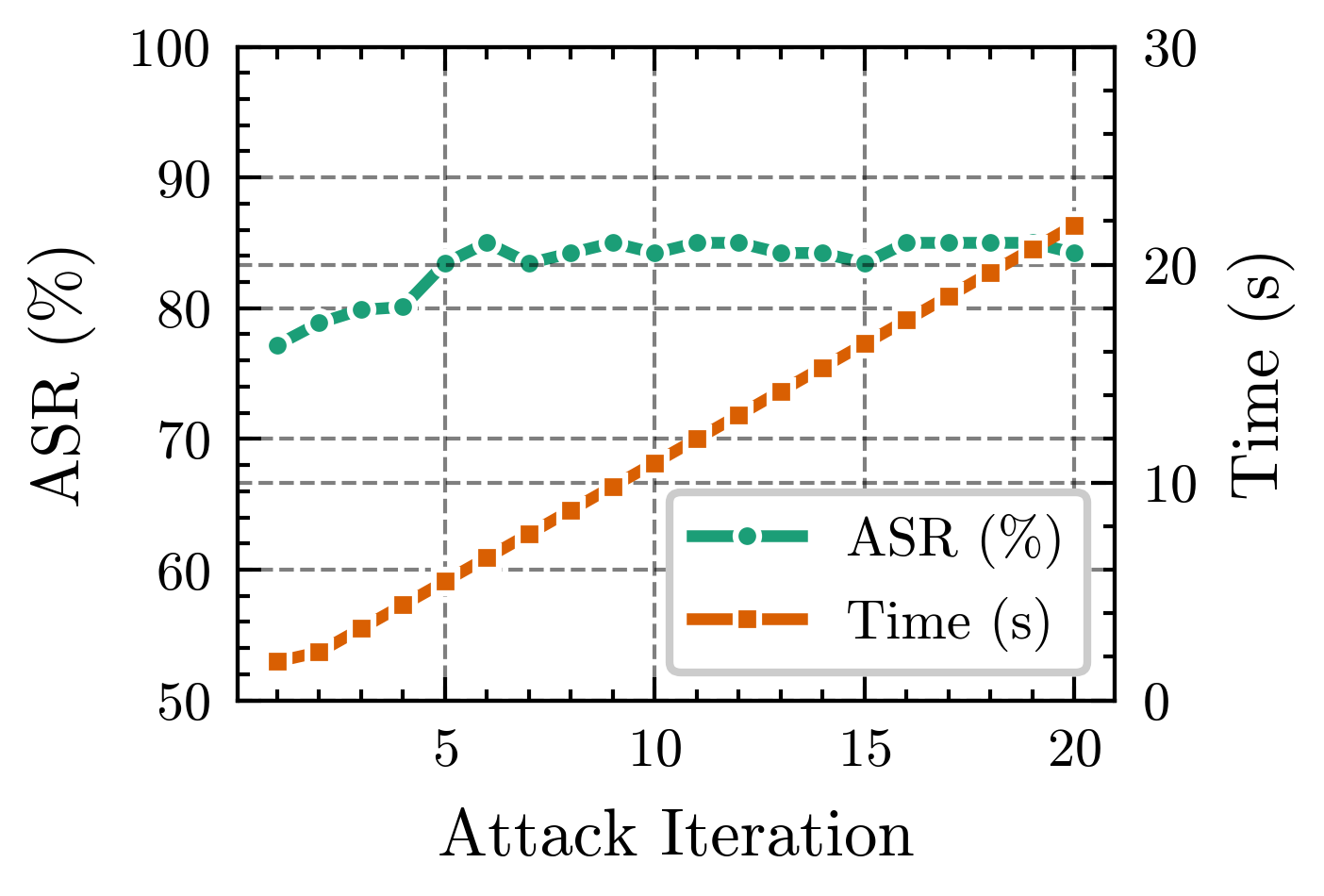}
    \caption{The ASRs and runtimes of \sysname over different attack iterations.}
    \label{eval_runtime}
\end{figure}

\revise{
We focus here on the practical costs associated with \sysname.
Generally, when conducting large-scale attacks in real-world scenarios, it is crucial to consider the runtime costs of attack methods.
The primary factors influencing the runtime of \sysname include the size of the proxy model and the number of attack iterations.
We demonstrate that using smaller models and fewer attack iterations allows effectively reducing the overall overhead of \sysname while maintaining competitive attack effectiveness.
}

\revise{
Figure \ref{eval_runtime} illustrates the ASRs and time costs of \sysname when using Res-18 as the proxy model across different attack iterations.
Here, we adjust the step size to ensure that the product of the step size and the number of attack iterations equals $\frac{16}{255}$.
Besides, we employ a batch size of 64 and use Inc-v4 as the target model.
We utilize a GTX 4090 GPU and report the average values over 10 trials.
We observe that when the number of attack iterations reaches 5, the ASRs stabilize around 85\%, with a runtime of approximately 4.9 seconds. Therefore, in practice, we can set the number of attack iterations to 5 to reduce the attack costs.
}

\revise{
Additionally, when using Res-152 with 10 iterations, the time cost is about 98.8 seconds.
As shown in Table \ref{result_normal_model}, although replacing Res-18 with Res-152 and using an attack iteration of 5 results in a decrease in the ASR by about 7 percentage points, the attack cost is reduced to one-fortieth of the original.
Furthermore, by employing mixed precision, we find that the runtime can be further reduced from 4.9 seconds to approximately 2.5 seconds, resulting in a cost that is one-eighty of the original.
We believe that generating adversarial examples for 64 samples in just 2.5 seconds is highly efficient in practice.
}

\section{Real-world Evaluation}

\begin{figure}[!t]
  \centering
  \begin{minipage}{.17\textwidth}
    \centering
    \includegraphics[width=0.85\textwidth]{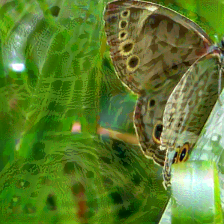}
  \end{minipage}%
  \begin{minipage}{.31\textwidth}
    \centering
    \vspace{-0.05cm}
    \raggedright
    \sloppy
    \textbf{Response:} This image presents an abstract and highly magnified view of what appears to be some kind of biological or organic structure. The image is dominated by intricate patterns, textures, and shapes in various shades of green, brown, and cream colors.
  \end{minipage}
    \vspace{+0.5cm}
    
  \begin{minipage}{.17\textwidth}
    \centering
    \includegraphics[width=.85\textwidth]{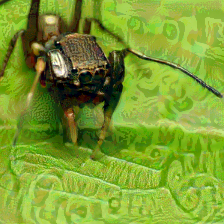}
  \end{minipage}%
  \begin{minipage}{.31\textwidth}
    \centering
    \vspace{-0cm}
    \raggedright
    \textbf{Response:} This image is a highly magnified and abstract view that appears to be of some kind of electronic circuit board or computer component. The surface is dominated by intricate patterns and textures in shades of green, brown, and metallic hues. At the center of the image, there is a rectangular structure that could be a microchip or integrated circuit.
  \end{minipage}
    \vspace{+0.5cm}
    
  \begin{minipage}{.17\textwidth}
    \centering
    \includegraphics[width=.85\textwidth]{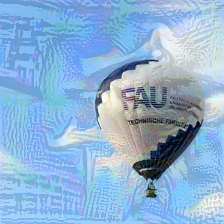}
  \end{minipage}%
  \begin{minipage}{.31\textwidth}
    \centering
    \vspace{-0cm}
    \raggedright
    \textbf{Response:} This image is a surreal and abstract digital composition that blends natural and technological elements in a thought-provoking way. At the center of the image is a realistic-looking hand, rendered in shades of gray and white. The hand appears to be grasping or holding something.
  \end{minipage} 
  \caption{The responses of Claude3 to the adversarial examples generated by \sysname are presented here. We prompted Claude3 with "Describe this image.".}
  \label{fig_claude}
\end{figure}

To conduct a more dependable and thorough evaluation of \sysname, we extend our evaluation to include attacks against Google MLaaS Classification System\footnote{https://cloud.google.com/vision/docs/drag-and-drop} on a multi-modal large-scale model Claude3\footnote{https://www.anthropic.com/claude} developed by AWS.
The Google MLaaS Classification System is recognized as one of the most advanced AI platforms available.
Claude3 ranks among the top-tier multi-modal models, according to the report from AWS\footnote{https://www.anthropic.com/news/claude-3-family}.
We employ Res-152 as the proxy model for this evaluation.
We randomly select 100 samples from the ImageNet and craft adversarial examples using BSR and \sysname.
It is important to highlight that the training dataset, model architecture, and task specifics of Res-152 are markedly different from those of Claude3.
This divergence serves as a stringent test to validate whether our attack can indeed effectively identify and exploit model-cross features within given images.

\subsection{Google MLaaS Classification System}

We record the Top-5 predictions for the generated adversarial examples and then enlist three volunteers to evaluate their effectiveness.
The evaluation protocol is to determine whether the entities in the adversarial examples appear in the corresponding Top-5 predictions.

The volunteers find that, for BSR, 57, 55, and 58 entities from the samples are included in the Top-5 predictions.
In contrast, \sysname yields 25, 26, and 25 entities in the Top-5 predictions.
These results indicate that the average ASRs for BSR and \sysname against Google Classification System are 43.33\% and 74.67\%, respectively, demonstrating the superior performance of \sysname.

Figure \ref{fig_google_attack} illustrates three successful attack examples.
While Google Vision System recognizes the background colors of three images, this information, however, is not particularly meaningful\footnote{During the assessment, the volunteers primarily focus on whether the entities in the images matched the given Top-5 predictions, rather than the background colors. As a result, all three samples are marked as successful attacks.}.
For instance, the top image depicts a butterfly, which is easily recognizable to the human eye.
However, Google Classification System misclassified this image with high confidence, identifying it as "Arthropod, Organism, Liquid."
This misclassification underscores the effectiveness of our adversarial examples in deceiving advanced AI systems.
\subsection{Claude3}

We feed adversarial examples produced by BSR and \sysname into Claude3, prompting it with "Describe this image".
Following this, we collect Claude3's responses and ask the same volunteers to assess the consistency between Claude3's responses and the semantic content of the generated adversarial examples, responding with either a yes or a no.
For BSR, the volunteers identify that 64, 68, and 60 of the samples, respectively, are semantically consistent, resulting in ASRs of 36\%, 32\%, and 40\%, respectively.
For \sysname, the volunteers find that 25, 29, and 18 of the samples, respectively, are semantically consistent, indicating ASRs of 75\%, 71\%, and 82\% for \sysname.
These ASRs further underscore the vulnerabilities of DNNs.

To provide a tangible understanding, Figure \ref{fig_claude} presents three illustrative examples.
Figure \ref{fig_claude} is divided into two columns: the left column displays the adversarial examples, while the right column showcases Claude3’s corresponding responses.
Due to space constraints, only the initial summary paragraph of Claude3's responses is presented.
Specifically, the top image in Figure \ref{fig_claude} shows an image of a butterfly, but Claude3 fails to recognize it.
More impressively, in the bottom images in Figure \ref{fig_claude}, Claude3 misidentifies an insect and a hot-air balloon as an electronic circuit board and a hand, respectively.
These misclassifications highlight the significant challenges DNNs face when confronted with adversarial examples, emphasizing the need for improved robustness in AI systems.

\section{Conclusion}
\label{conclusion}
In this paper, we studied the transferability of adversarial examples through the lens of Bayesian approach.
We answered how to combine the transferability of adversarial examples with Bayesian approach and identified what constitutes a transferability-promoting prior, so as to gain deeper insights into the underlying principles that govern the transferability.
Building upon this foundation, we designed two specific transferability-promoting priors, namely, pixel-level removal prior and region-based soft removal prior, accompanied by an adaptive dynamic weighting strategy.
We conducted extensive experiments to examine the performance of the proposed attack.
We also tested \sysname on a large-scale model deployed in a real-world scenario to further validate its effectiveness.
Looking forward, we hope that this study can alarm relevant parties about the potential vulnerabilities of DNNs and spur further development of sophisticated adversarial defense mechanisms.

\section{Acknowledgments}
This work was supported by the National Natural Science Foundation of China under grant number 62202170 and Alibaba Group through the Alibaba Innovation Research Program.

\clearpage

\bibliographystyle{ACM-Reference-Format}
\bibliography{sample-base}

\end{document}